\documentclass[11pt,a4paper]{article}
\usepackage[hyperref]{acl2019}
\usepackage{times}
\usepackage{latexsym}

\usepackage{url}
\usepackage{amssymb, amsmath, eqnarray}
\usepackage{times}
\usepackage{latexsym}
\usepackage{bm}
\usepackage{multirow}
\usepackage{graphicx}  %Required

\aclfinalcopy % Uncomment this line for the final submission
\def\aclpaperid{1637} %  Enter the acl Paper ID here

\newcommand*\myat{{\fontfamily{ptm}\selectfont\small @}}

\title{Bridging the Gap between Training and Inference \\for Neural Machine Translation}

\def\first{$^1$}
\def\second{$^2$}
\def\third{$^3$}
\def\fourth{$^4$}
\def\fifth{$^5$}
\def\comma{$^,$}
\author{Wen Zhang\first\comma\second ~~~ Yang Feng\first\comma\second\thanks{\ \ Corresponding author.} ~~~ Fandong Meng\third ~~~ Di You\fourth ~~~ Qun Liu\fifth
\\
\\
{ \first {Key Laboratory of Intelligent Information Processing}} \\ Institute of Computing Technology, Chinese Academy of Sciences (ICT/CAS) \\
{ \second {University of Chinese Academy of Sciences, Beijing, China}} \\
\texttt{\small\{\href{mailto:zhangwen@ict.ac.cn}{zhangwen},\href{mailto:fengyang@ict.ac.cn}{fengyang}\}\myat ict.ac.cn}\\
{ \third {Pattern Recognition Center, WeChat AI, Tencent Inc, China}}\\
\href{mailto:fandongmeng@tencent.com}{\texttt{\small fandongmeng\myat tencent.com}}\\
{ \fourth {Worcester Polytechnic Institute, Worcester, MA, USA}}\\
\href{mailto:diyou@gmail.com}{\texttt{\small dyou\myat wpi.edu}}\\
{ \fifth {Huawei Noah's Ark Lab, Hong Kong, China}}\\
\href{mailto:qun.liu@huawei.com}{\texttt{\small qun.liu\myat huawei.com}} \\
}

\date{}

\begin{document}

\maketitle
\begin{abstract}

Neural Machine Translation (NMT) generates target words sequentially in the way of predicting the next word conditioned on the context words. At training time, it predicts with the ground truth words as context while at inference it has to generate the entire sequence from scratch. This discrepancy of the fed context leads to error accumulation among the way. Furthermore, word-level training requires strict matching between the generated sequence and the ground truth sequence which leads to overcorrection over different but reasonable translations. In this paper, we address these issues by sampling context words not only from the ground truth sequence but also from the predicted sequence by the model during training, where the predicted sequence is selected with a sentence-level optimum. 
Experiment results on Chinese$\rightarrow$English and WMT'14 English$\rightarrow$German translation tasks demonstrate that our approach can achieve significant improvements on multiple datasets.

\end{abstract}

\section{Introduction} \label{sec:intro}

Neural Machine Translation has shown promising results and drawn more attention recently. Most NMT models fit in the encoder-decoder framework, including the RNN-based~\cite{sutskever2014sequence,bahdanau2014neural,meng2019dtmt}, the CNN-based~\cite{gehring2017convolutional} and the attention-based~\cite{ashish2017attention} models, which predict the next word conditioned on the previous context words, deriving a language model over target words. The scenario is at training time the ground truth words are used as context while at inference the entire sequence is generated by the resulting model on its own and hence the previous words generated by the model are fed as context. As a result, the predicted words at training and inference are drawn from different distributions, namely, from the data distribution as opposed to the model distribution. This discrepancy, called {\em exposure bias}~\cite{ranzato2015sequence}, leads to a gap between training and inference. As the target sequence grows, the errors accumulate among the sequence and the model has to predict under the condition it has never met at training time.

Intuitively, to address this problem, the model should be trained to predict under the same condition it will face at inference. Inspired by \textsc{Data As Demonstrator} (\textsc{DaD})~\cite{venkatraman2015improving}, feeding as context both ground truth words and the predicted words during training can be a solution. NMT models usually optimize the cross-entropy loss which requires a strict pairwise matching at the word level between the predicted sequence and the ground truth sequence. Once the model generates a word deviating from the ground truth sequence, the cross-entropy loss will correct the error immediately and draw the remaining generation back to the ground truth sequence. However, this causes a new problem. A sentence usually has multiple reasonable translations and it cannot be said that the model makes a mistake even if it generates a word different from the ground truth word. For example,

\begin{tabular}{ll}
\centering
{\em reference}: & We should comply with the rule. \\
{\em cand1}: & We should abide with the rule. \\
{\em cand2}: & We should abide by the law. \\
{\em cand3}: & We should abide by the rule. \\
\end{tabular}
once the model generates ``abide'' as the third target word, the cross-entropy loss would force the model to generate ``with'' as the fourth word (as {\em cand1}) so as to produce larger sentence-level likelihood and be in line with the reference, although ``by'' is the right choice. Then, ``with'' will be fed as context to generate ``the rule'', as a result, the model is taught to generate ``abide with the rule'' which actually is wrong. The translation {\em cand1} can be treated as {\em overcorrection} phenomenon. Another potential error is that even the model predicts the right word ``by'' following ``abide'', when generating subsequent translation, it may produce ``the law'' improperly by feeding ``by'' (as {\em cand2}). Assume the references and the training criterion let the model memorize the pattern of the phrase ``the rule'' always following the word ``with'', to help the model recover from the two kinds of errors and create the correct translation like {\em cand3}, we should feed ``with'' as context rather than ``by'' even when the previous predicted phrase is ``abide by''. We refer to this solution as {\em Overcorrection Recovery} ({\em OR}).

In this paper, we present a method to bridge the gap between training and inference and improve the overcorrection recovery capability of NMT. Our method first selects {\em oracle} words from its predicted words and then samples as context from the oracle words and ground truth words. Meanwhile, the oracle words are selected not only with a word-by-word greedy search but also with a sentence-level evaluation, e.g. BLEU, which allows greater flexibility under the pairwise matching restriction of cross-entropy. At the beginning of training, the model selects as context ground truth words at a greater probability. As the model converges gradually, oracle words are chosen as context more often. In this way, the training process changes from a fully guided scheme towards a less guided scheme. Under this mechanism, the model has the chance to learn to handle the mistakes made at inference and also has the ability to recover from overcorrection over alternative translations. We verify our approach on both the RNNsearch model and the stronger Transformer model. The results show that our approach can significantly improve the performance on both models.

\section{RNN-based NMT Model} \label{shallownmt}

Our method can be applied in a variety of NMT models. Without loss of generality, we take the RNN-based NMT~\cite{bahdanau2014neural} as an example to introduce our method. Assume the source sequence and the observed translation are $\bm{\mathrm{x}}=\{x_1,\cdots,x_{|\bm{\mathrm{x}}|}\}$ and $\bm{\mathrm{y}}^{*}=\{y_1^{*},\cdots,y_{|\bm{\mathrm{y}}^{*}|}^{*}\}$.

{\bf Encoder.}
A bidirectional Gated Recurrent Unit (GRU) \cite{cho2014learning} is used to acquire two sequences of hidden states, the annotation of $x_i$ is $h_i = [{\overrightarrow{h}_i};{\overleftarrow{h}_i}]$. Note that $e_{x_i}$ is employed to represent the embedding vector of the word $x_i$.
\begin{gather}
\overrightarrow{h}_i = \bm{\mathrm{GRU}}(e_{x_i}, \overrightarrow{h}_{i-1}) \label{eq:left} \\
\overleftarrow{h}_i =  \bm{\mathrm{GRU}}(e_{x_i}, \overleftarrow{h}_{i+1}) \label{eq:right}
\end{gather}

{\bf Attention.}
The attention is designed to extract source information (called source context vector). %which is highly related to the generation of the next target word.
At the $j$-th step, the relevance between the target word $y_j^{*}$ and the $i$-th source word is evaluated and normalized over the source sequence
\begin{gather} 
    r_{ij}=\bm{\mathrm{v}}_a^T \tanh\left(\bm{\mathrm{W}}_as_{j-1} + \bm{\mathrm{U}}_ah_i\right) \label{eq:att_query} \\
    \alpha_{ij} = \frac{\exp \left( r_{ij} \right)}{\sum_{i'=1}^{|\bm{\mathrm{x}}|} \exp \left( r_{i'j} \right)} \label{eq:att_alpha}
\end{gather}
The source context vector is the weighted sum of all source annotations and can be calculated by
\begin{equation} \label{eq:decode:c_j}
    c_j = \sum\nolimits_{i=1}^{|\bm{\mathrm{x}}|}\alpha_{ij}h_i
\end{equation}

{\bf Decoder.}
The decoder employs a variant of GRU to unroll the target information. At the $j$-th step, the target hidden state $s_j$ is given by
\begin{equation} \label{eq:decode:s}
    s_j = \bm{\mathrm{GRU}}(e_{y_{j-1}^{*}}, s_{j-1}, c_j)
\end{equation}
The probability distribution $P_{j}$ over all the words in the target vocabulary is produced conditioned on the embedding of the previous ground truth word, the source context vector and the hidden state
\begin{gather}
t_j = g\left(e_{y_{j-1}^{*}}, c_j, s_j\right)  \label{eq:t} \\
o_j = \bm{\mathrm{W}}_o t_j \label{eq:o} \\
P_{j} = \mathrm{softmax}\left(o_j\right)  \label{eq:softmax}
\end{gather}
where $g$ stands for a linear transformation, $\bm{\mathrm{W}}_o$ is used to map $t_j$ to $o_j$ so that each target word has one corresponding dimension in $o_j$.

\section{Approach}

\begin{figure}[!tb]
    \centering
    \includegraphics[scale=0.3]{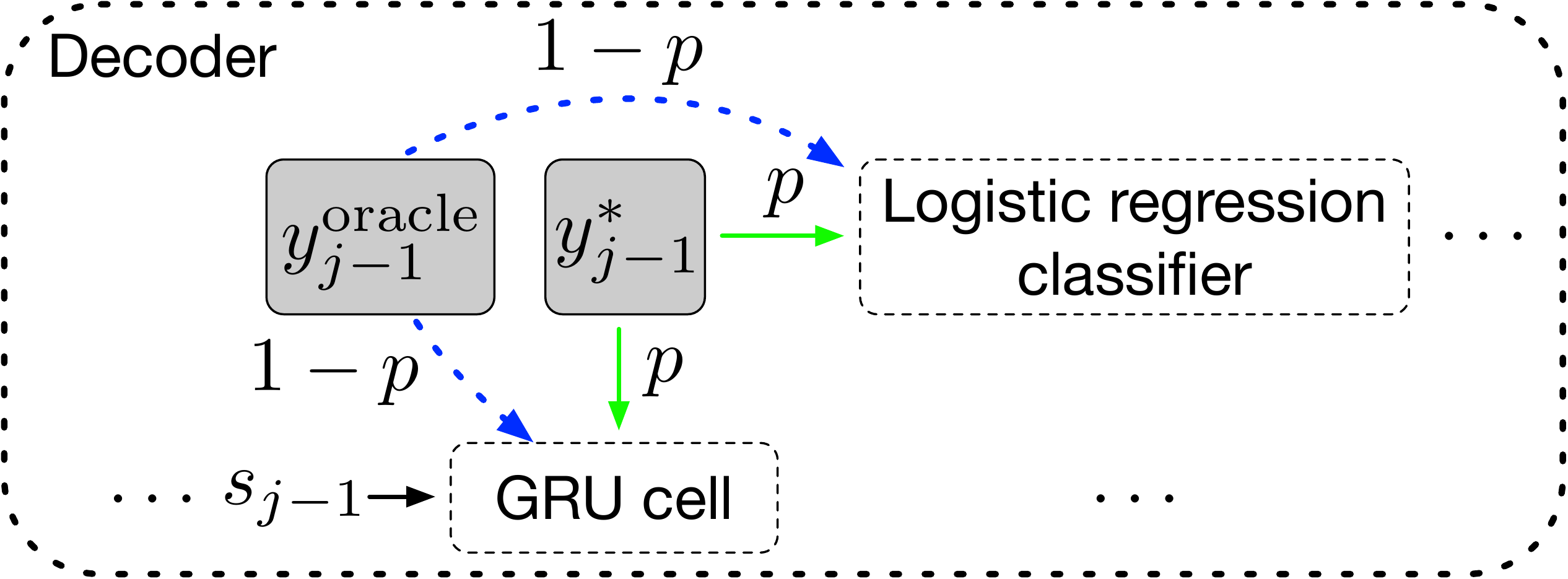}
    \caption{The architecture of our method.} 
    \label{fig:integration}
\end{figure}

The main framework (as shown in Figure~\ref{fig:integration}) of our method is to feed as context either the ground truth words or the previous predicted words, i.e. {\em oracle words}, with a certain probability. This potentially can reduce the gap between training and inference by training the model to handle the situation which will appear during test time. We will introduce two methods to select the oracle words. One method is to select the oracle words at the word level with a greedy search algorithm, and another is to select a oracle sequence at the sentence-level optimum. The sentence-level oracle provides an option of $n$-gram matching with the ground truth sequence and hence inherently has the ability of recovering from overcorrection for the alternative context. To predict the $j$-th target word $y_j$, the following steps are involved in our approach:

\begin{itemize}
\item[1.] Select an oracle word $y_{j-1}^\mathrm{oracle}$ (at word level or sentence level) at the \{$j$$-$$1$\}-th step. (Section {\bf Oracle Word Selection}) 
\item[2.] Sample from the ground truth word $y_{j-1}^*$ with a probability of $p$ or from the oracle word $y_{j-1}^\mathrm{oracle}$ with a probability of $1$$-$$p$. (Section {\bf Sampling with Decay}) 
\item[3.] Use the sampled word as $y_{j-1}$ and replace the $y_{j-1}^{*}$ in Equation (\ref{eq:decode:s}) and (\ref{eq:t}) with $y_{j-1}$, then perform the following prediction of the attention-based NMT.
\end{itemize}

%The whole procedure is shown in Figure~\ref{fig:integration}.

\subsection{Oracle Word Selection} \label{sec:oracle}
Generally, at the $j$-th step, the NMT model needs the ground truth word $y_{j-1}^{*}$ as the context word to predict $y_j$, thus, we could select an oracle word $y_{j-1}^\mathrm{oracle}$ to simulate the context word. The oracle word should be a word similar to the ground truth or a synonym. Using different strategies will produce a different oracle word $y_{j-1}^\mathrm{oracle}$. One option is that word-level greedy search could be employed to output the oracle word of each step, which is called {\em Word-level Oracle} (called $\mathrm{WO}$). Besides, we can further optimize the oracle by enlarging the search space with beam search and then re-ranking the candidate translations with a sentence-level metric, e.g. BLEU \cite{papineni2002bleu}, GLEU \cite{wu2016google}, ROUGE \cite{lin2004rouge}, etc, the selected translation is called {\em oracle sentence}, the words in the translation are {\em Sentence-level Oracle} (denoted as $\mathrm{SO}$).

\subsubsection*{Word-Level Oracle}

\begin{figure}[!bt]
    \centering
    \includegraphics[scale=0.27]{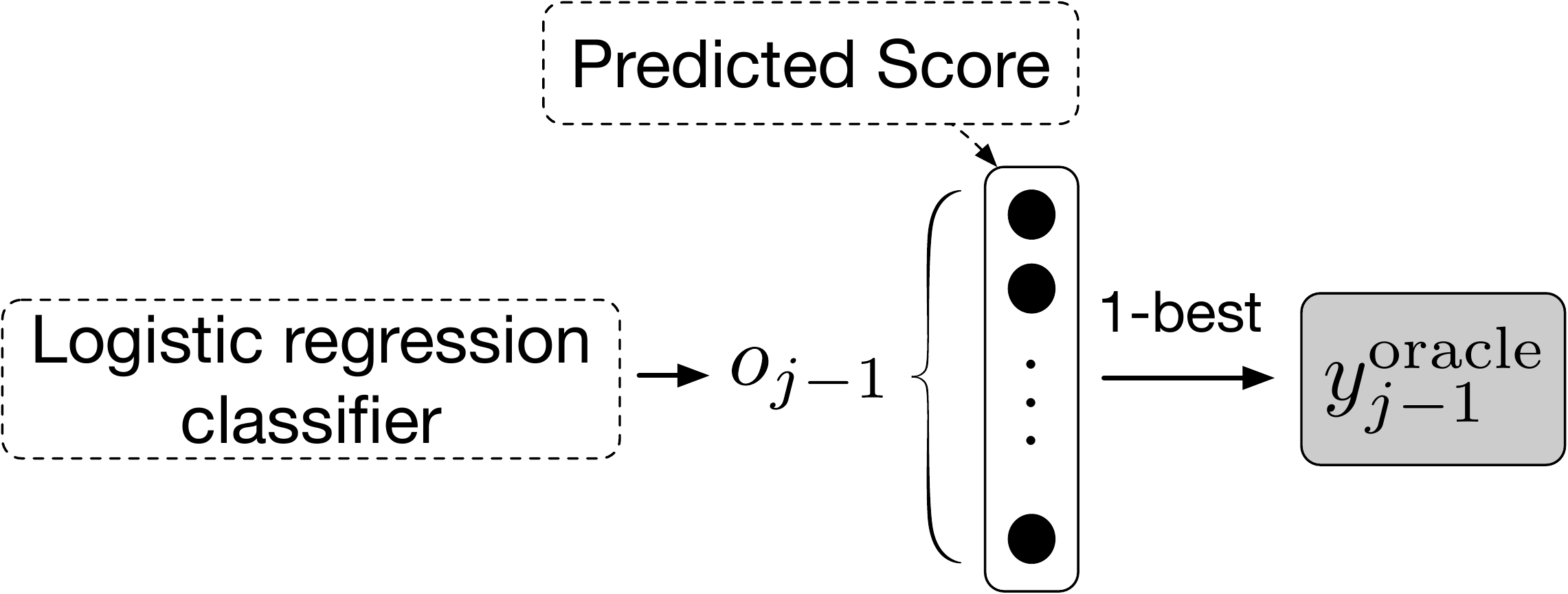}
    \caption{Word-level oracle without noise.}
    \label{fig:oracle_word}
\end{figure}

For the \{$j$$-$$1$\}-th decoding step, the direct way to select the word-level oracle is to pick the word with the highest probability from the word distribution $P_{j-1}$ drawn by Equation (\ref{eq:softmax}), which is shown in Figure~\ref{fig:oracle_word}. The predicted score in $o_{j-1}$ is the value before the $\mathrm{softmax}$ operation. In practice, we can acquire more robust word-level oracles by introducing the {\em Gumbel-Max} technique \cite{gumbel1954statistical,maddison2014Asampling}, which provides a simple and efficient way to sample from a categorical distribution.

\begin{figure}[!bt]
    \centering
    \includegraphics[scale=0.27]{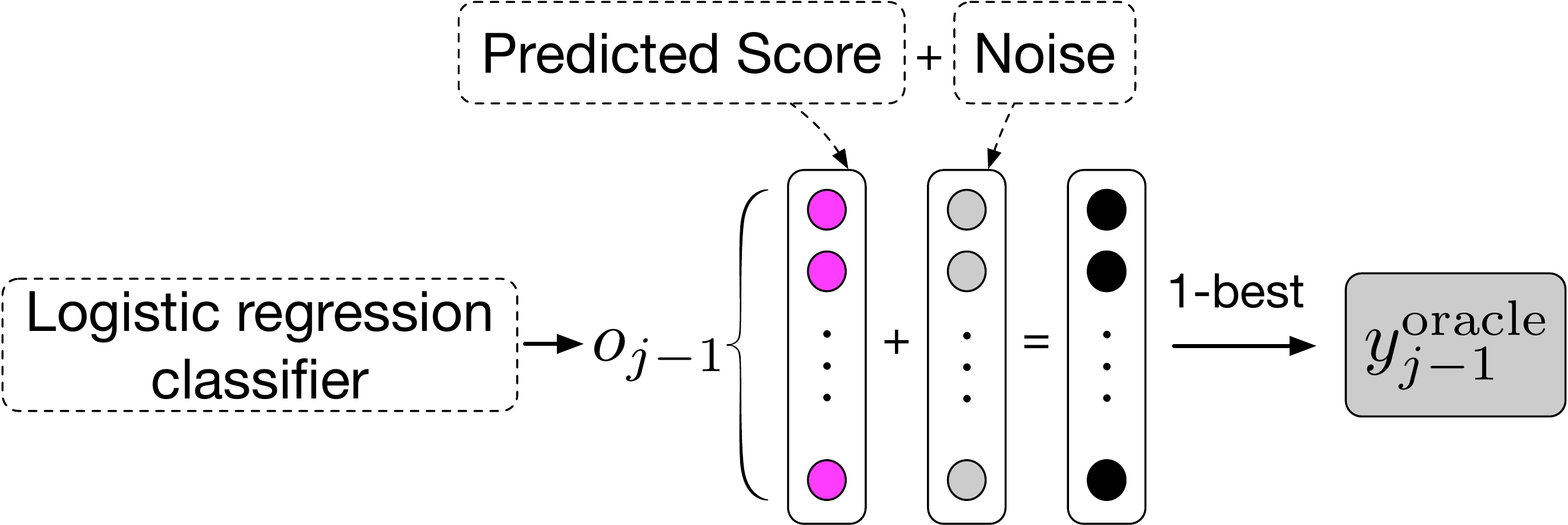}
    \caption{Word-level oracle with Gumbel noise.} 
    \label{fig:oracle-gumbel}
\end{figure}

The Gumbel noise, treated as a form of regularization, is added to $o_{j-1}$ in Equation (\ref{eq:o}), as shown in Figure~\ref{fig:oracle-gumbel}, then $\mathrm{softmax}$ function is performed, the word distribution of $y_{j-1}$ is approximated by
\begin{gather}
\eta=-\log\left(-\log u\right) \label{eq:gumbel} \\
\tilde{o}_{j-1} = \left(o_{j-1} + \eta\right) / \tau \label{eq:gumbel-o} \\
\tilde{P}_{j-1} = \mathrm{softmax}\left(\tilde{o}_{j-1}\right)  \label{eq:gumbel-softmax}
\end{gather}
where $\eta$ is the Gumbel noise calculated from a uniform random variable $u\sim\mathcal{U}(0,1)$, $\tau$ is temperature. As $\tau$ approaches 0, the $\mathrm{softmax}$ function is similar to the $\mathrm{argmax}$ operation, and it becomes uniform distribution gradually when $\tau \rightarrow \infty$. Similarly, according to $\tilde{P}_{j-1}$, the $1$-best word is selected as the word-level oracle word
\begin{equation}
y_{j-1}^\mathrm{oracle} = y_{j-1}^\mathrm{WO} = \mathrm{argmax}\left(\tilde{P}_{j-1}\right)
\end{equation}
Note that the Gumbel noise is just used to select the oracle and it does not affect the loss function for training.

\subsubsection*{Sentence-Level Oracle}

The sentence-level oracle is employed to allow for more flexible translation with $n$-gram matching required by a sentence-level metric. In this paper, we employ BLEU as the sentence-level metric. To select the sentence-level oracles, we first perform beam search for all sentences in each batch, assuming beam size is $k$, and get $k$-best candidate translations. In the process of beam search, we also could apply the Gumbel noise for each word generation. We then evaluate each translation by calculating its BLEU score with the ground truth sequence, and use the translation with the highest BLEU score as the {\em oracle sentence}. We denote it as $\bm{\mathrm{y}}^\mathrm{S}=(y_1^\mathrm{S}, ..., y_{|\bm{\mathrm{y}}^\mathrm{S}|}^\mathrm{S})$, then at the $j$-th decoding step, we define the sentence-level oracle word as
\begin{equation} \label{f_oracle_2}
    y_{j-1}^\mathrm{oracle} = y_{j-1}^\mathrm{SO} = y_{j-1}^\mathrm{S}  \\
\end{equation}
But a problem comes with sentence-level oracle. As the model samples from ground truth word and the sentence-level oracle word at each step, the two sequences should have the same number of words. However we can not assure this with the naive beam search decoding algorithm. Based on the above problem, we introduce {\em force decoding} to make sure the two sequences have the same length.

\textbf{Force Decoding.} As the length of the ground truth sequence is $|\bm{\mathrm{y}}^{*}|$, the goal of force decoding is to generate a sequence with $|\bm{\mathrm{y}}^{*}|$ words followed by a special end-of-sentence (EOS) symbol. Therefore, in beam search, once a candidate translation tends to end with EOS when it is shorter or longer than $|\bm{\mathrm{y}}^{*}|$, we will force it to generate $|\bm{\mathrm{y}}^{*}|$ words, that is,
\begin{itemize}
\item If the candidate translation gets a word distribution $P_{j}$ at the $j$-th step where $j \leqslant |\bm{\mathrm{y}}^{*}|$ and EOS is the top first word in $P_{j}$, then we select the top second word in $P_{j}$ as the $j$-th word of this candidate translation.
\item If the candidate translation gets a word distribution $P_{|\bm{\mathrm{y}}^{*}|+1}$ at the \{$|\bm{\mathrm{y}}^{*}|$$+$$1$\}-th step where EOS is not the top first word in $P_{|\bm{\mathrm{y}}^{*}|+1}$, then we select EOS as the \{$|\bm{\mathrm{y}}^{*}|$$+$$1$\}-th word of this candidate translation.
\end{itemize}
In this way, we can make sure that all the $k$ candidate translations have $|\bm{\mathrm{y}}^{*}|$ words, then re-rank the $k$ candidates according to BLEU score and select the top first as the oracle sentence.
For adding Gumbel noise into the sentence-level oracle selection, we replace the $P_{j}$ with $\tilde{P}_{j}$ at the $j$-th decoding step during force decoding.

\subsection{Sampling with Decay} \label{sec:sample}

In our method, we employ a sampling mechanism to randomly select the ground truth word $y^*_{j-1}$ or the oracle word $y^\mathrm{oracle}_{j-1}$ as $y_{j-1}$. At the beginning of training, as the model is not well trained, using $y^\mathrm{oracle}_{j-1}$ as $y_{j-1}$ too often would lead to very slow convergence, even being trapped into local optimum. On the other hand, at the end of training, if the context $y_{j-1}$ is still selected from the ground truth word $y^*_{j-1}$ at a large probability, the model is not fully exposed to the circumstance which it has to confront at inference and hence can not know how to act in the situation at inference. In this sense, the probability $p$ of selecting from the ground truth word can not be fixed, but has to decrease progressively as the training advances. At the beginning, $p$$=$$1$, which means the model is trained entirely based on the ground truth words. As the model converges gradually, the model selects from the oracle words more often.

Borrowing ideas from but being different from~\citet{bengio2015scheduled} which used a schedule to decrease $p$ as a function of the index of mini-batch, we define $p$ with a decay function dependent on the index of training epochs $e$ (starting from $0$)
\begin{equation} \label{eq:decay}
    p=\frac{\mu}{\mu+\exp\left(e/\mu\right)}
\end{equation}
where $\mu$ is a hyper-parameter. The function is strictly monotone decreasing. As the training proceeds, the probability $p$ of feeding ground truth words decreases gradually.

\subsection{Training}

After selecting $y_{j-1}$ by using the above method, we can get the word distribution of $y_j$ according to Equation (\ref{eq:decode:s}), (\ref{eq:t}), (\ref{eq:o}) and (\ref{eq:softmax}). We do not add the Gumbel noise to the distribution when calculating loss for training. The objective is to maximize the probability of the ground truth sequence based on maximum likelihood estimation (MLE). Thus following loss function is minimized:
\begin{equation} \label{f_loss}
    \mathcal{L}\left(\theta\right) = -\sum\nolimits_{n=1}^{N}\sum\nolimits_{j=1}^{|\bm{\mathrm{y}}^n|}\log P_{j}^n\left[y_j^n\right]
\end{equation}
where $N$ is the number of sentence pairs in the training data, $|\bm{\mathrm{y}}^n|$ indicates the length of the $n$-th ground truth sentence, $P_{j}^n$ refers to the predicted probability distribution at the $j$-th step for the $n$-th sentence, hence $P_{j}^n\left[y_j^n\right]$ is the probability of generating the ground truth word $y_j^n$ at the $j$-th step.

\section{Related Work}

Some other researchers have noticed the problem of exposure bias in NMT and tried to solve it.~\citet{venkatraman2015improving} proposed \textsc{Data As Demonstrator} (DAD) which initialized the training examples as the paired two adjacent ground truth words and at each step added the predicted word paired with the next ground truth word as a new training example.~\citet{bengio2015scheduled} further developed the method by sampling as context from the previous ground truth word and the previous predicted word with a changing probability, not treating them equally in the whole training process. This is similar to our method, but they do not include the sentence-level oracle to relieve the overcorrection problem and neither the noise perturbations on the predicted distribution.

Another direction of attempts is the sentence-level training with the thinking that the sentence-level metric, e.g., BLEU, brings a certain degree of flexibility for generation and hence is more robust to mitigate the exposure bias problem.
To avoid the problem of exposure bias,~\citet{ranzato2015sequence} presented a novel algorithm Mixed Incremental Cross-Entropy Reinforce (MIXER) for sequence-level training, which directly optimized the sentence-level BLEU used at inference.~\citet{shen2016minimum} introduced the Minimum Risk Training (MRT) into the end-to-end NMT model, which optimized model parameters by minimizing directly the expected loss with respect to arbitrary evaluation metrics, e.g., sentence-level BLEU.~\citet{shao2018greedy} proposed to eliminate the exposure bias through a probabilistic n-gram matching objective, which trains NMT NMT under the greedy decoding strategy.
%the method achieved significant improvements over MLE.

\section{Experiments}
We carry out experiments on the NIST Chinese$\rightarrow$English (Zh$\rightarrow$En) and the WMT'14 English$\rightarrow$German (En$\rightarrow$De) translation tasks.

\subsection{Settings}
For Zh$\rightarrow$En, the training dataset consists of 1.25M sentence pairs extracted from LDC corpora\footnote{These sentence pairs are mainly extracted from LDC2002E18, LDC2003E07, LDC2003E14, Hansards portion of LDC2004T07, LDC2004T08 and LDC2005T06}. We choose the NIST 2002 (MT02) dataset as the validation set, which has $878$ sentences, and the NIST 2003 (MT03), NIST 2004 (MT04), NIST 2005 (MT05) and NIST 2006 (MT06) datasets as the test sets, which contain 919, 1788, 1082 and 1664 sentences respectively.
For En$\rightarrow$De, we perform our experiments on the corpus provided by WMT'14, which contains 4.5M sentence pairs\footnote{\url{http://www.statmt.org/wmt14/translation-task.html}}. We use the $\mathrm{newstest2013}$ as the validation set, and the $\mathrm{newstest2014}$ as the test sets, which containing $3003$ and $2737$ sentences respectively. 
We measure the translation quality with BLEU scores~\cite{papineni2002bleu}. For Zh$\rightarrow$En, case-insensitive BLEU score is calculated by using the \emph{ mteval-v11b.pl} script. For En$\rightarrow$De, we tokenize the references and evaluate the performance with case-sensitive BLEU score by the \emph{ multi-bleu.pl} script. The metrics are exactly the same as in previous work. Besides, we make statistical significance test according to the method of~\citet{collins2005clause}.

%\begin{table}[ht!]
%\centering
%\renewcommand\arraystretch{1.2}
%\begin{tabular}{l||c|c}
%{\bf Task }  & {\bf Source} & {\bf Target} \\ \hline
%Zh$\rightarrow$En   & $30,000$ & $30,000$  \\ \hline
%En$\xrightarrow{16k}$De   & $19,272$ & $19,002$ \\
%\end{tabular}
%\caption{Statistics of vocabulary sizes on the two training datasets. Zh$\rightarrow$En means no BPE was applied. $16k$ is the number of merging operations for BPE.}
%\label{tb:vocab}
%\end{table}

\begin{table*}[ht!]
\centering
\renewcommand\arraystretch{0.9}
\begin{tabular}{l|l||c|c|c|c|c}
{\bf Systems }  & {\bf Architecture} & {\bf MT03} & {\bf MT04} & {\bf MT05} & {\bf MT06} & {\bf Average} \\ \hline
\multicolumn{7}{c}{\em Existing end-to-end NMT systems} \\ \hline
%\citeauthor{bahdanau2014neural}~\shortcite{bahdanau2014neural}		& RNNsearch  		  & 33.70  & 36.15     & 31.81 	& 32.71 	& 33.59 \\
~\citet{Tu2016}		        & Coverage	   	  & $33.69$	& $38.05$	& $35.01$	 & $34.83$	& $35.40$\\
~\citet{shen2016minimum} 	& MRT  		  	  & $37.41$  & $39.87$     & $37.45$	& $36.80$	& $37.88$ \\
~\citet{zhangEtalACL2017} 	& Distortion		  & $37.93$	& $40.40$	& $36.81$	&   $35.77$	& $37.73$\\	
%\citeauthor{wangEtAl2017}~\shortcite{wangEtAl2017}  				& DeepLAU	  	  & 39.35	& 41.15	& 38.07	& 37.29	& 38.97 \\
%\citeauthor{MENGIJCAI2018}~\shortcite{MENGIJCAI2018}  			& KVMemory	  	  & 38.40	& 41.10	& 38.73	& 39.08	& 39.33 \\
%\citeauthor{shen2016minimum}~\shortcite{shen2016minimum} 		& MRT  		  	  & $34.50$ & $36.95$ & $34.46$ & $33.24$ & {\em $34.79$}  \\ %Adadelta
%\citeauthor{mengEtAlCOLING2016}~\shortcite{mengEtAlCOLING2016}	& \textsc{MemAtt}	  & 35.69	& 39.24	& 35.74	 & 35.10	& 36.44\\
%\citeauthor{WangLLL16}~\shortcite{WangLLL16}  					& \textsc{MemDec}	  & 36.16	& 39.81	& 35.91	& 35.98	& 36.95\\
\hline
\multicolumn{7}{c}{\em Our end-to-end NMT systems} \\ \hline
\multirow{7}{*}{this work}  			&  RNNsearch		& $37.93$ & $40.53$ & $36.65$  & $35.80$ & $37.73$ \\
    						&   ~~~+ SS-NMT   	& $38.82$ & $41.68$ & $37.28$ & $37.98$ & $38.94$  \\
						&   ~~~+ MIXER      	& $38.70$ & $40.81$ & $37.59$ & $38.38$ & $38.87$  \\
						&   ~~~+ OR-NMT      	& {\bf 40.40$^{\ddag\dag\star}$} & {\bf 42.63$^{\ddag\dag\star}$} & {\bf 38.87$^{\ddag\dag\star}$} & {\bf 38.44$^{\ddag}$} & {\bf 40.09} \\
\cline{2-7}
%sentence oracle   & 39.22 & 42.11 & 38.33 & 38.59 & {\em 39.56} & - \\ \hline
                        &  Transformer		& $46.89$ & $47.88$ & $47.40$ & $46.66$ & $47.21$ \\
    					&  ~~~+ word oracle       & $47.42$  & $48.34$ & $47.89$ & $47.34$ & $47.75$\\ 
    					&  ~~~+ sentence oracle   & {\bf 48.31$^{\ast}$} & {\bf 49.40$^{\ast}$} & {\bf 48.72$^{\ast}$} & {\bf 48.45$^{\ast}$} & {\bf 48.72}\\ 
\hline
\end{tabular}
\caption{Case-insensitive BLEU scores (\%) on Zh$\rightarrow$En translation task. ``$\ddag$", ``$\dag$", ``$\star$" and ``$\ast$" indicate statistically significant difference (p\textless0.01) from RNNsearch, SS-NMT, MIXER and Transformer, respectively.}
\label{tb:compare_zh_en}
\end{table*}

In training the NMT model, we limit the source and target vocabulary to the most frequent $30$K words for both sides in the Zh$\rightarrow$En translation task, covering approximately $97.7$\% and $99.3$\% words of two corpus respectively. 
For the En$\rightarrow$De translation task, sentences are encoded using byte-pair encoding (BPE)~\cite{sennrich2015neural} with $37k$ merging operations for both source and target languages, which have vocabularies of $39418$ and $40274$ tokens respectively. We limit the length of sentences in the training datasets to $50$ words for Zh$\rightarrow$En and $128$ subwords for En$\rightarrow$De.
For RNNSearch model, the dimension of word embedding and hidden layer is $512$, and the beam size in testing is $10$. All parameters are initialized by the uniform distribution over $\left[-0.1,0.1\right]$. The mini-batch stochastic gradient descent (SGD) algorithm is employed to train the model parameters with batch size setting to $80$. Moreover, the learning rate is adjusted by adadelta optimizer~\cite{zeiler2012adadelta} with $\rho$=$0.95$ and $\epsilon$=$1e\textnormal{-}6$. Dropout is applied on the output layer with dropout rate being $0.5$.
For Transformer model, we train base model with default settings (fairseq\footnote{\url{https://github.com/pytorch/fairseq}}).

\subsection{Systems}
The following systems are involved:
%{\bf RNNsearch.}
%We implemented the attention-based NMT of ~\citeauthor{bahdanau2014neural}~\shortcite{bahdanau2014neural} with PyTorch deep learning framework\footnote{\url{http://pytorch.org}}.

\paragraph{RNNsearch:} Our implementation of an improved model as described in Section~\ref{shallownmt}, where the decoder employs two GRUs and an attention. Specifically, Equation~\ref{eq:decode:s} is substituted with:
\begin{gather}
\tilde{s}_j = \bm{\mathrm{GRU}}_1(e_{y_{j-1}^{*}}, s_{j-1}) \label{eq:improve_dec_0} \\
s_j = \bm{\mathrm{GRU}}_2(c_j, \tilde{s}_j)  \label{eq:improve_dec_1}
\end{gather}
Besides, in Equation~\ref{eq:att_query}, $s_{j-1}$ is replaced with $\tilde{s}_{j-1}$.

\paragraph{SS-NMT:} Our implementation of the scheduled sampling (SS) method~\cite{bengio2015scheduled} on the basis of the RNNsearch. The decay scheme is the same as Equation~\ref{eq:decay} in our approach.

\paragraph{MIXER:} Our implementation of the mixed incremental cross-entropy reinforce~\cite{ranzato2015sequence}, where the sentence-level metric is BLEU and the average reward is acquired according to its offline method with a $1$-layer linear regressor.

\paragraph{OR-NMT:} Based on the RNNsearch, we introduced the word-level oracles, sentence-level oracles and the Gumbel noises to enhance the overcorrection recovery capacity. For the sentence-level oracle selection, we set the beam size to be $3$, set $\tau$=$0.5$ in Equation (\ref{eq:gumbel-o}) and $\mu$=$12$ for the decay function in Equation (\ref{eq:decay}). OR-NMT is the abbreviation of NMT with Overcorrection Recovery.

\subsection{Results on Zh$\rightarrow$En Translation}
We verify our method on two baseline models with the NIST Zh$\rightarrow$En datasets in this section.

\subsubsection*{Results on the RNNsearch}
As shown in Table~\ref{tb:compare_zh_en},~\citet{Tu2016} propose to model coverage in RNN-based NMT to improve the adequacy of translations.~\citet{shen2016minimum} propose minimum risk training (MRT) for NMT to directly optimize model parameters with respect to BLEU scores.~\citet{zhangEtalACL2017} model distortion to enhance the attention model. Compared with them, our baseline system RNNsearch 1) outperforms previous shallow RNN-based NMT system equipped with the coverage model~\cite{Tu2016}; and 2) achieves competitive performance with the MRT~\cite{shen2016minimum} and the Distortion~\cite{zhangEtalACL2017} on the same datasets. We hope that the strong shallow baseline system used in this work makes the evaluation convincing.

We also compare with the other two related methods that aim at solving the exposure bias problem, including the scheduled sampling~\cite{bengio2015scheduled} (SS-NMT) and the sentence-level training~\cite{ranzato2015sequence} (MIXER). From Table~\ref{tb:compare_zh_en}, we can see that both SS-NMT and MIXER can achieve improvements by taking measures to mitigate the exposure bias. While our approach OR-NMT can outperform the baseline system RNNsearch and the competitive comparison systems by directly incorporate the sentence-level oracle and noise perturbations for relieving the overcorrection problem. Particularly, our OR-NMT significantly outperforms the RNNsearch by +$2.36$ BLEU points averagely on four test datasets. Comparing with the two related models, our approach further gives a significant improvements on most test sets and achieves improvement by about +$1.2$ BLEU points on average.

\subsubsection*{Results on the Transformer}
The methods we propose can also be adapted to the stronger Transformer model. The evaluated results are listed in Table~\ref{tb:compare_zh_en}. Our word-level method can improve the base model by +$0.54$ BLEU points on average, and the sentence-level method can further bring in +$1.0$ BLEU points improvement.

\begin{table}[!t]
\centering
\renewcommand\arraystretch{1.}
\begin{tabular}{l||c}
{\bf Systems }  & {\bf Average} \\
\hline
%RNNsearch    	& $33.59$  \\ \cline{2-2}
RNNsearch   & $37.73$ \\ \hline
	~~~+ word oracle   & $38.94$ \\ 
	~~~~~~~~+ noise  & $39.50$ \\ 
	~~~+ sentence oracle   & $39.56$  \\ 
	~~~~~~~~+ noise  & {\bf 40.09} \\
\end{tabular}
\caption{Factor analysis on Zh$\rightarrow$En translation, the results are average BLEU scores on MT03$\sim$06 datasets.}
\label{tb:factors}
\end{table}

\subsection{Factor Analysis}
We propose several strategies to improve the performance of approach on relieving the overcorrection problem, including utilizing the word-level oracle, the sentence-level oracle, and incorporating the Gumbel noise for oracle selection. To investigate the influence of these factors, we conduct the experiments and list the results in Table~\ref{tb:factors}.

%We first employed the word-level oracle, then changed to the sentence-level oracle, and last added the Gumbel noise to make the word-level and sentence-level word selection. The context word $y_{j-1}$ is always sampled with decay.
%The results are given in Table~\ref{tb:factors}. Note that the RNNsearch$^{\star}$ with word-level oracle is exactly the same with SS-NMT system.
When only employing the word-level oracle, the translation performance was improved by +$1.21$ BLEU points, this indicates that feeding predicted words as context can mitigate exposure bias. When employing the sentence-level oracle, we can further achieve +$0.62$ BLEU points improvement. It shows that the sentence-level oracle performs better than the word-level oracle in terms of BLEU. We conjecture that the superiority may come from a greater flexibility for word generation which can mitigate the problem of overcorrection. By incorporating the Gumbel noise during the generation of the word-level and sentence-level oracle words, the BLEU score are further improved by $0.56$ and $0.53$ respectively. This indicates Gumbel noise can help the selection of each oracle word, which is consistent with our claim that Gumbel-Max provides a efficient and robust way to sample from a categorical distribution.
\begin{figure}[tbp]
    \centering
    \includegraphics[scale=0.5]{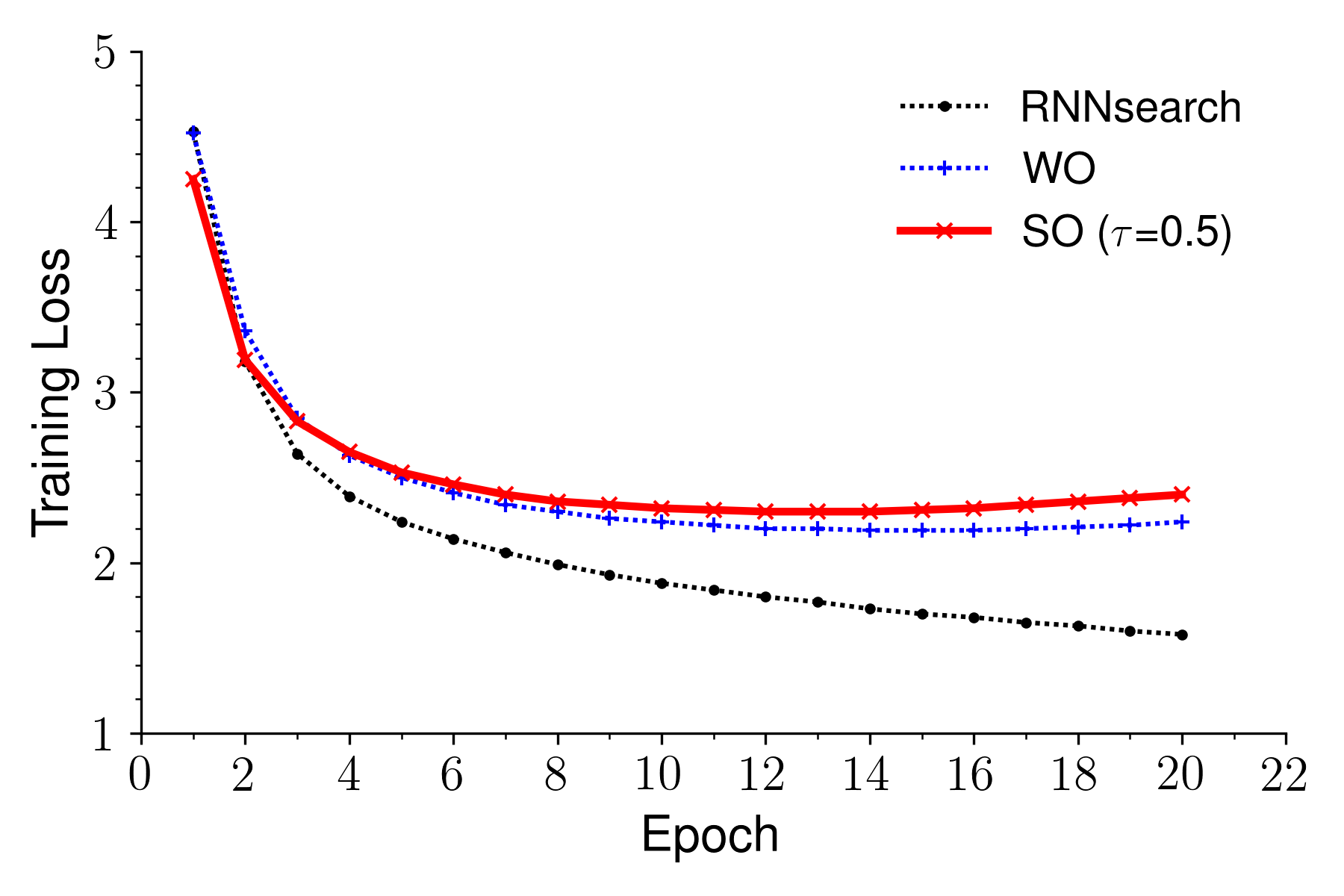}
    \caption{Training loss curves on Zh$\rightarrow$En translation with different factors. The black, blue and red colors represent the RNNsearch, RNNsearch with word-level oracle and RNNsearch with sentence-level oracle systems respectively.}
    \label{fig:training_curve_loss}
\end{figure}
\subsection{About Convergence}

\begin{figure}[!t]
    \centering
    \includegraphics[scale=0.5]{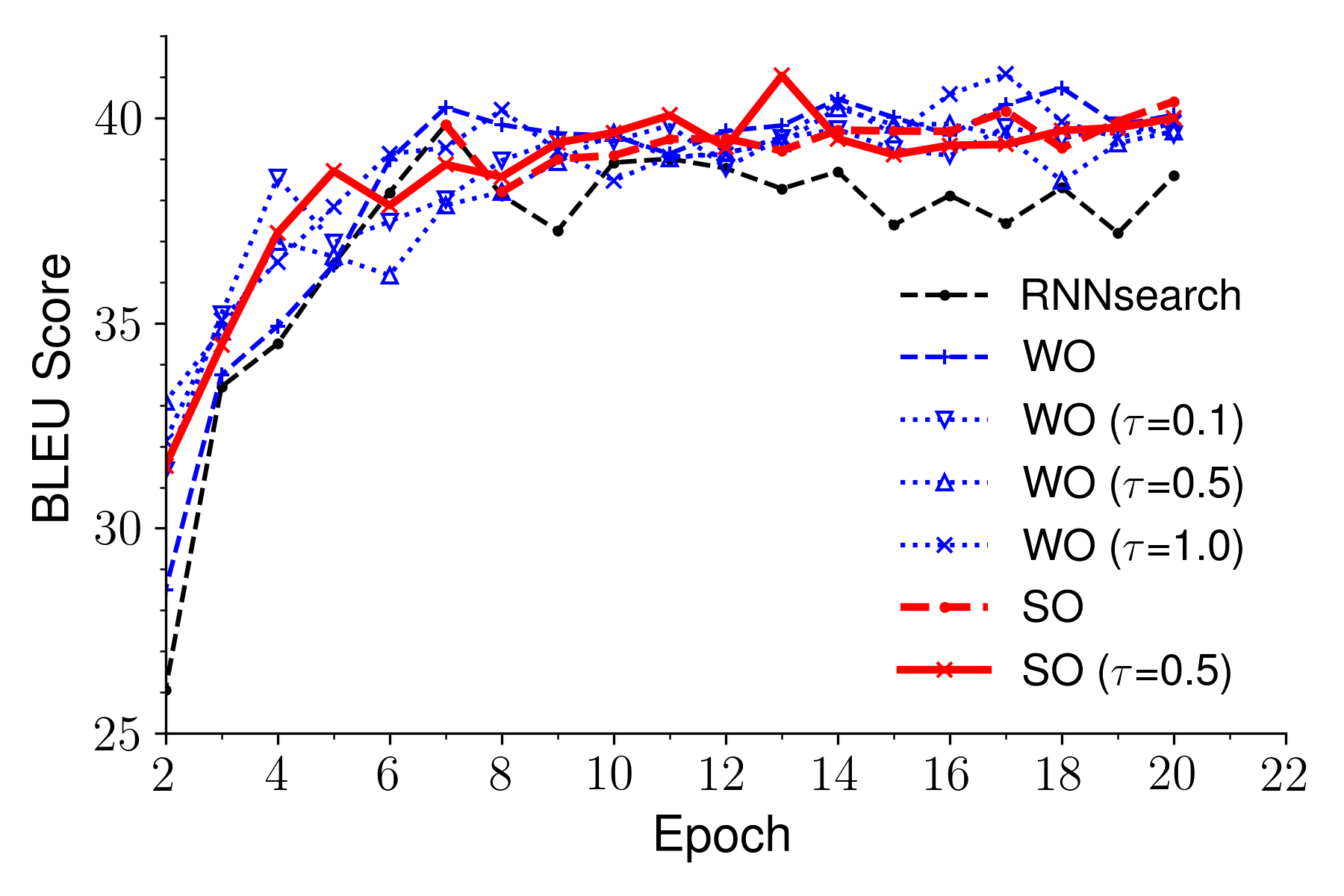}
    \caption{Trends of BLEU scores on the validation set with different factors on the Zh$\rightarrow$En translation task.}
    \label{fig:dev_curve_bleu}
\end{figure}

In this section, we analyze the influence of different factors for the convergence. Figure~\ref{fig:training_curve_loss} gives the training loss curves of the RNNsearch, word-level oracle (WO) without noise and sentence-level oracle (SO) with noise. In training, BLEU score on the validation set is used to select the best model, a detailed comparison among the BLEU score curves under different factors is shown in Figure~\ref{fig:dev_curve_bleu}. RNNsearch converges fast and achieves the best result at the $7$-th epoch, while the training loss continues to decline after the $7$-th epoch until the end. Thus, the training of RNNsearch may encounter the overfitting problem.
%We first used the word-level and sentence-level oracle without the Gumbel noise, then included it in the system with different $\tau$.
%According to the characteristics of the Gumbel noise, when $\tau$ approaches $0$, the Gumbel-Softmax is similar to the argmax operation, and it approaches uniform distribution gradually when $\tau \rightarrow \infty$.
%Curves of the word oracle and sentence oracle with noise in 
Figure~\ref{fig:training_curve_loss} and~\ref{fig:dev_curve_bleu} also reveal that, integrating the oracle sampling and the Gumbel noise leads to a little slower convergence and the training loss does not keep decreasing after the best results appear on the validation set. This is consistent with our intuition that oracle sampling and noises can avoid overfitting despite needs a longer time to converge. %Interestingly, the proposed method causes the training loss to rise a little bit after the $14$-th epoch. 

\begin{figure}[!t]
    \centering
    \includegraphics[scale=0.5]{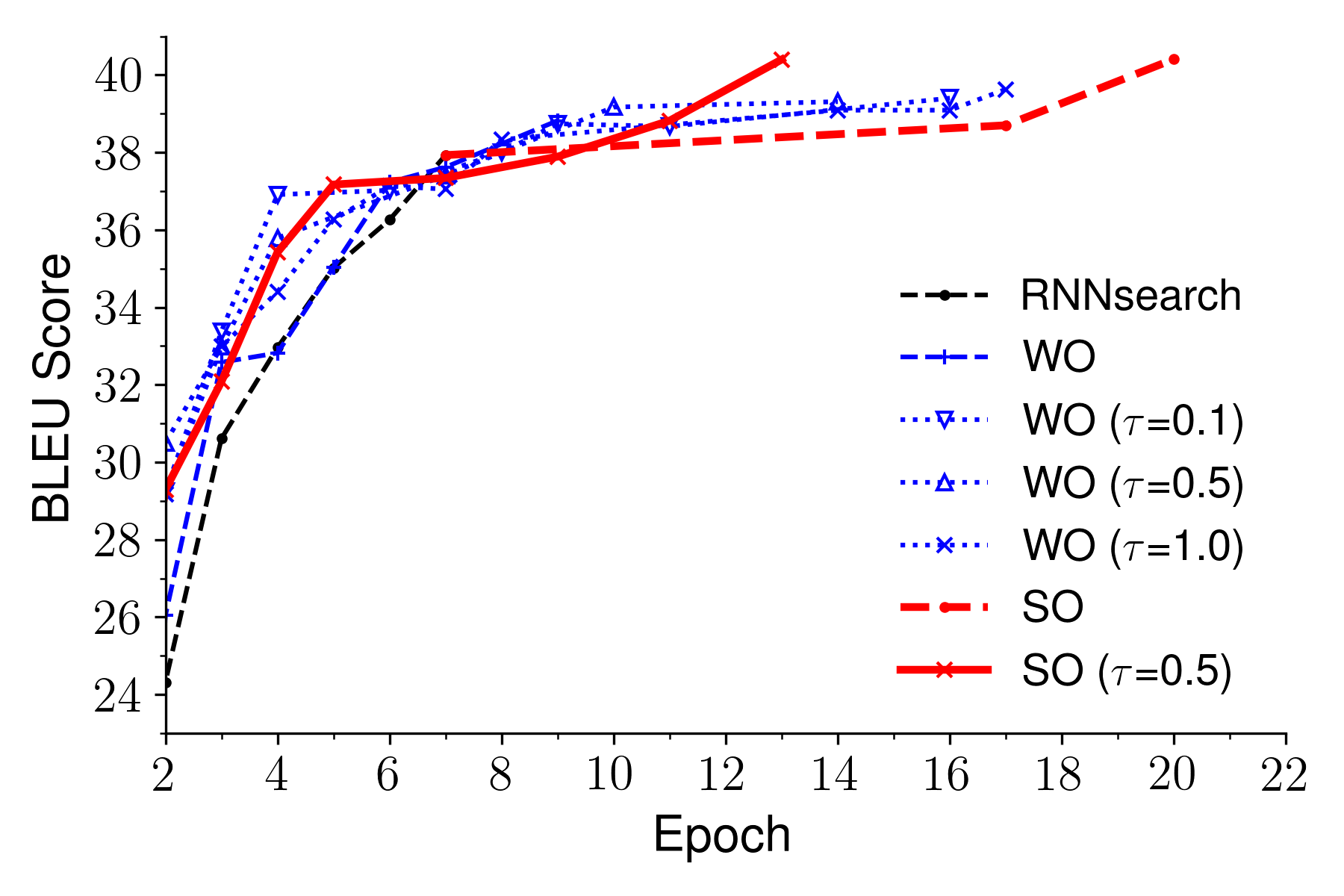}
    \caption{Trends of BLEU scores on the MT03 test set with different factors on the Zh$\rightarrow$En translation task.}
    \label{fig:tau_affect_nist03}
\end{figure}

Figure~\ref{fig:tau_affect_nist03} shows the BLEU scores curves on the MT03 test set under different factors\footnote{Note that the ``SO" model without noise is trained based on the pre-trained RNNsearch model (as shown by the red dashed lines in Figure~\ref{fig:dev_curve_bleu} and~\ref{fig:tau_affect_nist03}).}. When sampling oracles with noise ($\tau$=$0.5$) on the sentence level, we obtain the best model.
%If $\tau=0.1$, it is too close to the original distribution while $\tau=1.0$, the noise is far from the original distribution.
Without noise, our system converges to a lower BLEU score. This can be understood easily that using its own results repeatedly during training without any regularization will lead to overfitting and quick convergence. In this sense, our method benefits from the sentence-level sampling and Gumbel noise.
%the Gumbel noise as a form of regularization is effective in our method.

\subsection{About Length}

\begin{figure}[!t]
    \centering
    \includegraphics[scale=0.23]{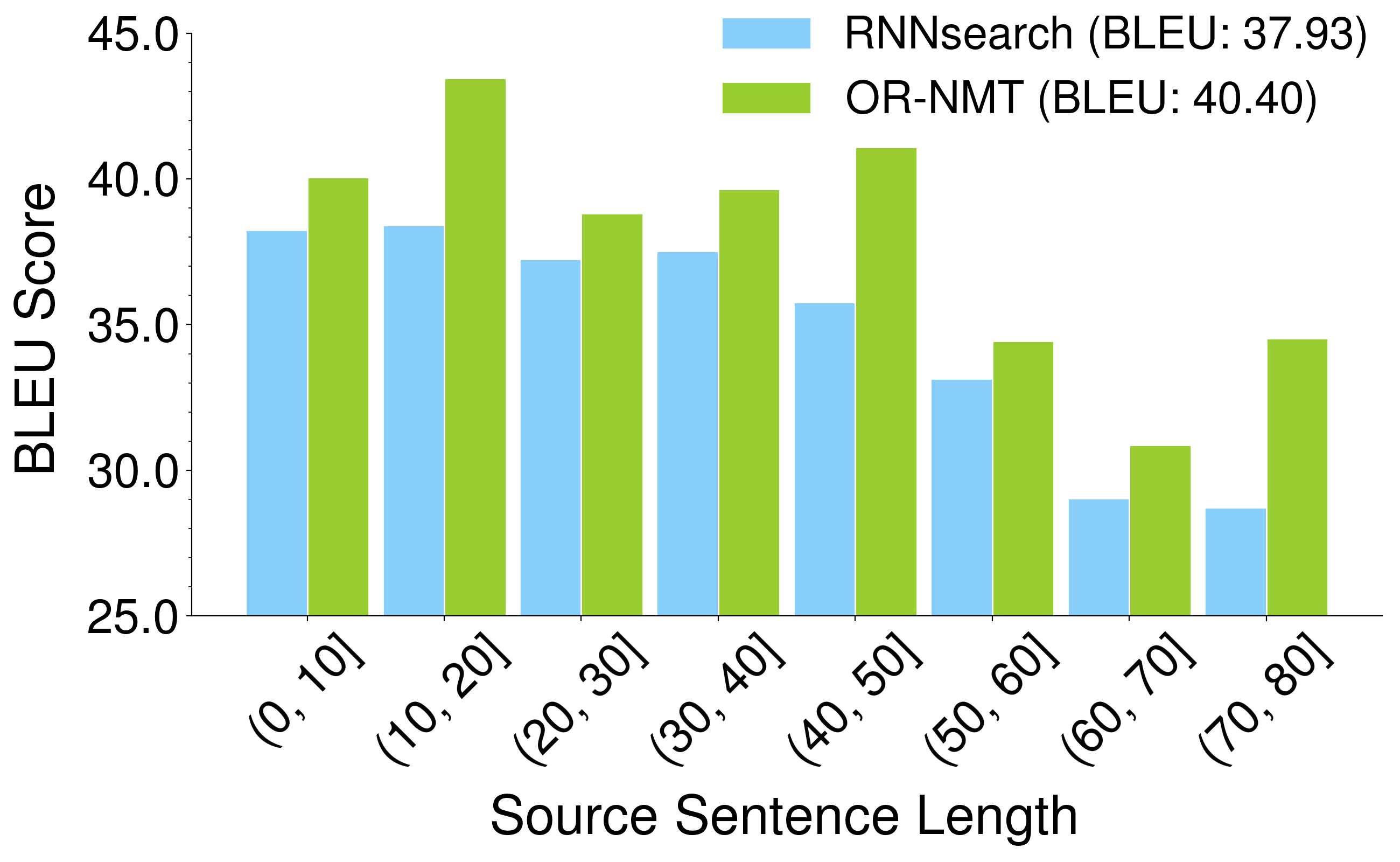}
    \caption{Performance comparison on the MT03 test set with respect to the different lengths of source sentences on the Zh$\rightarrow$En translation task.}
    \label{fig:len}
\end{figure}

Figure~\ref{fig:len} shows the BLEU scores of generated translations on the MT03 test set with respect to the lengths of the source sentences. In particular, we split the translations for the MT03 test set into different bins according to the length of source sentences, then test the BLEU scores for translations in each bin separately with the results reported in Figure~\ref{fig:len}. Our approach can achieve big improvements over the baseline system in all bins, especially in the bins ($10$,$20$], ($40$,$50$] and ($70$,$80$] of the super-long sentences. The cross-entropy loss requires that the predicted sequence is exactly the same as the ground truth sequence which is more difficult to achieve for long sentences, while our sentence-level oracle can help recover from this kind of overcorrection.
\subsection{Effect on Exposure Bias}
To validate whether the improvements is mainly obtained by addressing the exposure bias problem, we randomly select $1$K sentence pairs from the Zh$\rightarrow$En training data, and use the pre-trained RNNSearch model and proposed model to decode the source sentences. The BLEU score of RNNSearch model was $24.87$, while our model produced +$2.18$ points. We then count the ground truth words whose probabilities in the predicted distributions produced by our model are greater than those produced by the baseline model, and mark the number as $\mathcal{N}$. There are totally $28,266$ gold words in the references, and $\mathcal{N}$=$18,391$. The proportion is $18,391/28,266$=$65.06\%$, which could verify the improvements are mainly obtained by addressing the exposure bias problem.

\subsection{Results on En$\rightarrow$De Translation}

\begin{table}[!htb]
\centering
\renewcommand\arraystretch{1.0}
\begin{tabular}{l||c}
{\bf Systems}         & $\mathrm{newstest2014}$ \\ \hline
 RNNsearch    	      & $25.82$   \\
 ~~~+ SS-NMT          & $26.50$   \\ 
 ~~~+ MIXER           & $26.76$  \\ 
 ~~~+ OR-NMT    	  & {\bf 27.41$^{\ddag}$}  \\ \hline
  Transformer (base)  & $27.34$   \\
 ~~~+ SS-NMT          & $28.05$   \\ 
 ~~~+ MIXER           & $27.98$  \\ 
 ~~~+ OR-NMT    	  & {\bf 28.65$^{\ddag}$}  \\
\end{tabular}
\caption{Case-sensitive BLEU scores (\%) on En$\rightarrow$De task. The ``$\ddag$" indicates the results are significantly better (p\textless0.01) than RNNsearch and Transformer.}
\label{tb:compare_en_de}
\end{table}
We also evaluate our approach on the WMT'14 benchmarks on the En$\rightarrow$De translation task. From the results listed in Table~\ref{tb:compare_en_de}, we conclude that the proposed method significantly outperforms the competitive baseline model as well as related approaches. Similar with results on the Zh$\rightarrow$En task, both scheduled sampling and MIXER could improve the two baseline systems. Our method improves the RNNSearch and Transformer baseline models by +$1.59$ and +$1.31$ BLEU points respectively. These results demonstrate that our model works well across different language pairs.

\section{Conclusion}
The end-to-end NMT model generates a translation word by word with the ground truth words as context at training time as opposed to the previous words generated by the model as context at inference. To mitigate the discrepancy between training and inference, when predicting one word, we feed as context either the ground truth word or the previous predicted word with a sampling scheme. The predicted words, %used as context%
referred to as oracle words, can be generated with the word-level or sentence-level optimization. Compared to word-level oracle, sentence-level oracle can further equip the model with the ability of overcorrection recovery. To make the model fully exposed to the circumstance at reference, we sample the context word with decay from the ground truth words. 
We verified the effectiveness of our method with two strong baseline models and related works on the real translation tasks, achieved significant improvement on all the datasets. We also conclude that the sentence-level oracle show superiority over the word-level oracle.

\section*{Acknowledgments}

We thank the three anonymous reviewers for their valuable suggestions. 
This work was supported by National Natural Science Foundation of China (NO. 61662077, NO. 61876174) and National Key R\&D Program of China (NO. YS2017YFGH001428).

\bibliography{acl2019}
\bibliographystyle{acl_natbib}

\end{document}